\documentclass{article}

\usepackage{arxiv}

\usepackage[utf8]{inputenc} 
\usepackage[T1]{fontenc}    
\usepackage{hyperref}       
\usepackage{url}            
\usepackage{booktabs}       
\usepackage{amsfonts}       
\usepackage{nicefrac}       
\usepackage{microtype}      
\usepackage{lipsum}
\usepackage{bbm}            
\usepackage{float}          

\usepackage{amsmath}
\usepackage{setspace}
\usepackage{mathtools}
\usepackage{accents}
\usepackage{amssymb}
\usepackage{enumitem}
\usepackage{hyperref}
\usepackage{enumitem}
\usepackage{longtable}
\usepackage{fixmath}
\usepackage{wrapfig}
\usepackage{color}
\usepackage{algorithm}
\usepackage{algpseudocode}
\usepackage{comment}
\usepackage{subfig}
\usepackage{soul}
\usepackage{lineno}
\usepackage{graphicx,multicol}
\usepackage{url}
\usepackage{hyperref}
\usepackage{multicol,multirow}
\usepackage{booktabs}
\DeclareMathOperator*{\argmax}{arg\,max}
\DeclareMathOperator*{\argmin}{arg\,min}

\newcommand{\US}{$U\!\!S$} 

\title{Gradient-boosted based structured and unstructured learning}

\author{
  Andrea Trevi\~no Gavito\\
  Department of Industrial Engineering and Management Sciences\\
  Northwestern University\\
  Evanston, IL 60208 \\
  \texttt{andrea.tg@u.northwestern.edu} \\
   \And
   Diego Klabjan \\
  Department of Industrial Engineering and Management Sciences\\
  Northwestern University\\
  Evanston, IL 60208 \\
  \texttt{d-klabjan@northwestern.edu} \\
   \And
   Jean Utke \\ 
  Data, Discovery and Decision Science \\
  Allstate Insurance Company\\
  Northbrook, IL 60062 \\
  \texttt{jutke@allstate.com} \\
}

\begin{document}
\maketitle

\begin{abstract}
We propose two frameworks to deal with problem settings in which both structured and unstructured data are available. Structured data problems are best solved by traditional machine learning models such as boosting and tree-based algorithms, whereas deep learning has been widely applied to problems dealing with images, text, audio, and other unstructured data sources. However, for the setting in which both structured and unstructured data are accessible, it is not obvious what the best modeling approach is to enhance performance on both data sources simultaneously. Our proposed frameworks allow joint learning on both kinds of data by integrating the paradigms of boosting models and deep neural networks. The first framework, the boosted-feature-vector deep learning network, learns features from the structured data using gradient boosting and combines them with embeddings from unstructured data via a two-branch deep neural network. Secondly, the two-weak-learner boosting framework extends the boosting paradigm to the setting with two input data sources. We present and compare first- and second-order methods of this framework. Our experimental results on both public and real-world datasets show performance gains achieved by the frameworks over selected baselines by magnitudes of 0.1\% - 4.7\%.
    
\end{abstract}

\section{Introduction}


    Data in modern machine learning problems can be represented by a variety of modalities or data sources. We consider the setting in which structured data, aka tabular data, and unstructured data are available simultaneously. A common application area of this setting is medical diagnosis, in which the decision making process is supported by unstructured data such as medical imaging and doctors' notes, in combination with patient historical data, lab analyses and blood tests, in the form of structured data. 
    
    The settings where structured or unstructured data are available individually have been extensively researched. Deep neural networks (DNNs) have consistently proven successful at solving problems with unstructured data, see e.g. textbooks \cite{book1} and \cite{Goodfellow-et-al-2016}. On the other hand, traditional boosting methods have shown significant advantages over DNNs in modeling structured data inputs \cite{survey, fttransf2021, Caruana:2008, Caruana:2006,fttransf2021}. Examples of such benefits are observed in terms of training time, interpretability, amount of required training data, tuning efforts, and computational expense. This is commonly observed in Kaggle competitions, where better performance is achieved by boosted methods when the available data is structured \cite{RePEc:eee,DBLP:MangalK17,Taieb13agradient}, and by deep learning models when the available data is unstructured \cite{ Graham, DBLP:ZouXL17}. In particular, LightGBM \cite{NIPS2017_6907} and XGBoost \cite{Chen:2016}, have become de-facto modeling standards for structured data.
    
    Conversely, in the setting in which both structured and unstructured data are accessible (\US), it is not obvious what the best modeling approach is to enhance performance on both data sources simultaneously. In general, the simplest method consists of training independent models for each data modality and then combining the results by averaging or voting over the individual predictions. A big caveat is the missed opportunity of capturing any cross-data source interactions or underlying complementary information that might exist in the data. Training concurrently on both modalities of data is deemed crucial if we attempt to learn such relationships. A common approach to joint training consists of using DNNs for representation learning on each data source, concatenating the learned embeddings, and having it as input to a third DNN. This approach serves as a baseline for our experiments, and performs sub-optimally given that boosting algorithms excel on structured data settings.
    
    In this paper, we propose two frameworks for the \US setting that address the above-mentioned considerations. Our frameworks aim at better capturing the best and most informative features of each data source, while simultaneously enhancing performance though a joint training scheme. To achieve this, our novel approaches combine the proven paradigms for structured and unstructured data respectively: gradient boosting machines and DNNs. 
    
    The first framework is the boosted-feature-vector deep learning network (BFV+DNN). BFV+DNN learns features from the structured data using gradient boosting and combines them with embeddings from unstructured data via a two-branch deep neural network. It requires to train a boosted model on the structured data as an initial step. Then, each neural network branch learns embeddings specific to each data input, which are further fused into a shared trainable model. The post-fusion shared architecture allows the model to learn the complementary cross-data source interactions. The key novelty is the feature extraction process from boosting. Following standard terminology, we refer to model inputs as "features" and to DNN-learned representations as "embeddings." In our proposed framework, BFVs are used as inputs to BFV+DNN and hence, are named features accordingly.
    
    In addition, we propose a two-weak-learner boosting framework (2WL) that extends the boosting paradigm to the \US setting. The framework is derived as a first-order approximation to the gradient boosting risk function and further expanded to a second-order approximation method (2WL2O). It should be noted that this framework can be used in the general multimodal setting and is not restricted to the  \US use case. Our experimental results show significant performance gains over the aforementioned baseline. Relative improvements on F1 metrics are observed by magnitudes of 4.7\%, 0.1\%, and 0.34\% on modified Census, Imagenet, and Covertype datasets, respectively. We also consider a real-world dataset from an industry partner where the improvement in accuracy is 0.41\%.
    
    The main contributions of this work are as follows. 
    \begin{enumerate}
    \item We present a boosted-feature-vector DNN model that combines structured data boosting features with deep neural networks to address the setting in which both structured and unstructured data sources are available. 
    \item We propose an alternative two-weak-learner-gradient-boosting framework to address the setting in which both structured and unstructured data sources are available. 
    \item We extend the two-weak-learner-gradient-boosting to a second-order approximation. 
    \item We show and compare the effectiveness of these approaches on public and real-world datasets.
    \end{enumerate}

    The rest of this paper is organized as follows. In Section \ref{sec:lr}, we discuss related work. In Section \ref{sec:model}, we formally introduce our proposed frameworks. Experimental results are discussed in Section \ref{sec:comp} and we conclude in Section \ref{sec:conclusion}.

\section{Related work} \label{sec:lr}
\subsection{Boosting Methods}
    Boosting methods combine base models (referred to as weak learners) as a means to improve the performance achieved by individual learners \cite{ridgeway:1999}. AdaBoost \cite{Freund:1997} is one of the first concrete adaptive boosting algorithms, whereas Gradient Boosting Machines (GBM) \cite{Friedman00greedyfunction} derive the boosting algorithm from the perspective of optimizing a loss function using gradient descent, see \cite{ mayr:2014}. A formulation of gradient boosting for the multi-class setting and two algorithmic approaches are proposed in \cite{NIPS2011_4450}. LightGBM \cite{NIPS2017_6907} incorporates techniques to improve GBM's efficiency and scalability. Traditionally, trees have been the base learners of choice for boosting methods, but the performance of neural networks as weak learners for AdaBoost has also been investigated in \cite{Schwenk:2000}. More recently, CNNs were explored as weak learners for GBM in \cite{BMVC2016}, integrating the benefits of boosting algorithms with the impressive results that CNNs have obtained at learning representations on visual data,  \cite{DBLP:journals/corr/JiaSDKLGGD14, NIPS2012_4824, RedmonDGF16}. Second-order information is employed in boosting algorithms such as Logitboost \cite{Friedman98}, Taylorboost  \cite{SaberianMV11}, and XGBoost \cite{Chen:2016}. However, unlike our proposed second-order model, all these algorithms consider a single family of weak learners and individual data inputs, whereas we handle two families of weak learners and both structured and unstructured data simultaneously.

    Boosting approaches have also been applied to the setting in which more than one data source is available as input. For instance, a multiview boosting algorithm based on PAC-Bayesian theory is presented in \cite{Goyal:2018}  and a cost-based multimodal approach that introduces the notion of weak and strong modalities in \cite{KocoCB12}. In both cases, final classification is performed using majority or weighted voting. Similarly, a model that assigns a different contribution of each data input to the final classification is proposed in \cite{PengASP18}, where a shared weight distribution among modalities is used. None of these algorithms make use of DNN approaches, as they employ traditional decision stumps as weak learners regardless of the data input sources. In contrast, the multimodal reward-penalty-based voting boosting model proposed in \cite{LahiriPB18} uses DNNs as weak learners, but overlooks the benefits of tree-based approaches for structured data. Common to all methods reviewed in this paragraph, is the notion of addressing the setting where multiple data inputs are available. However, they do not take into account the underlying properties of these different data sources, nor do they consider specific algorithmic approaches that better suit each one of them.

\subsection{Structured \& Unstructured Data Setting}
 Approaches that directly target the \US setting are scarce, more so those that address the structured data characteristics. In \cite{Chen:2017}, the authors deal with demographics, living habits, and examination results from patients in the form of structured data, and with doctor's records and patient's medical history presented as unstructured text data. An intuitive DNN approach is used, with the drawbacks that have already been discussed as no special treatment is given to structured data.
 
 Conversely, in \cite{48133} the \US setting is tackled by combining the benefits of tree-based models and DNNs. To do so, they use stacking and boosted stacking of independently trained models. The core idea is similar to our proposed boosted-feature-vector DNN in the sense that a first model is trained and then used as an input for joint training. Their approaches differ from our BFV+DNN in two main aspects. First, their models are heavily tailored for the learning-to-rank use case and second, they use direct outputs from the first model as input to the second model, whereas we propose a novel way to extract boosted-feature vectors from the first model, rather than using its direct output.
 
\section{Proposed models}  \label{sec:model}

In this section, we propose two models to tackle the \US setting. The models address the inherent nature of each source of data by exploiting the specific benefits of boosted algorithms and neural networks as learners on each of them. 

\subsection{ Boosted-feature-vector Deep Learning Network (BFV+DNN)}
\begin{figure} [h]
    \centering
    \subfloat[Baseline architecture]{%
      \includegraphics[width=0.3\linewidth]{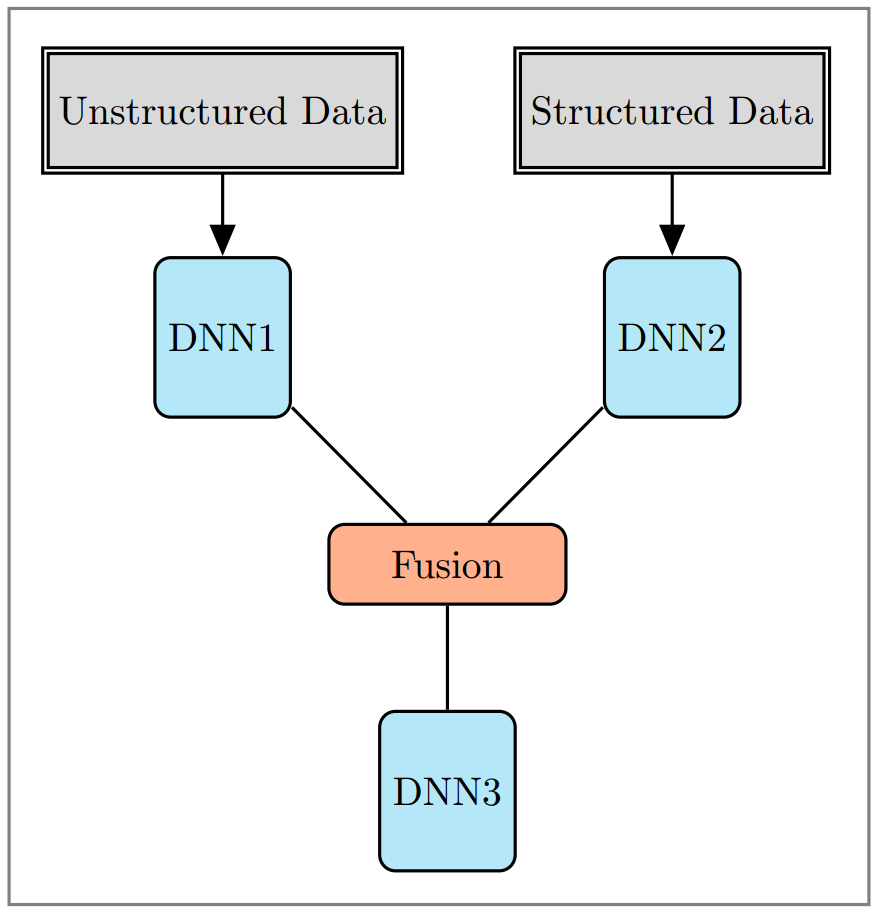}}
    \subfloat[BFV+DNN architecture]{%
      \includegraphics[width=0.3255\linewidth]{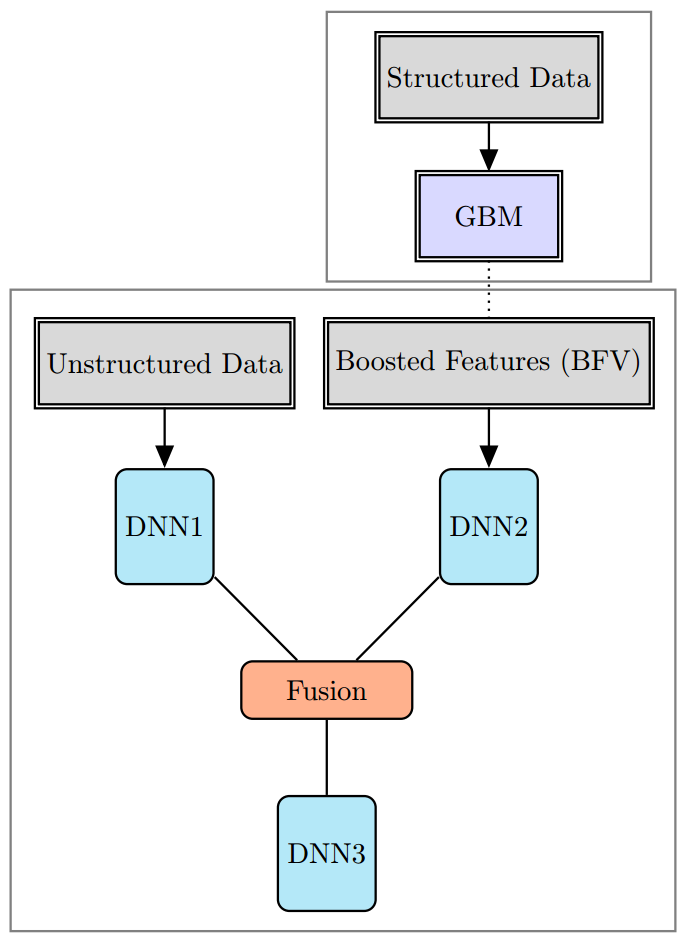}}
    \caption{\US deep neural networks} \label{exp:dnns} 
    \end{figure}

    The boosted-feature-vector deep learning network aims at using DNNs as the primary learning method, while incorporating boosted-feature vectors (BFV) from the structured data source.
    As a means of comparison, the baseline DNN approach to the \US setting is shown in Figure \ref{exp:dnns}a and the BFV+DNN architecture in Figure \ref{exp:dnns}b. Both contain two branches, a fusion stage and a joint learning architecture. Each branch learns representations from one data source (see DNN1 and DNN2 in Figure \ref{exp:dnns}). Then, a fusion yields a joint embedding that combines the data-source-specific representations. Finally, the combined vector is used as input to a trainable DNN to model cross-data source interactions (DNN3 in Figure \ref{exp:dnns}).

    As noted, the common baseline ignores the structured or unstructured nature of the data source and directly learns representations via DNNs, whereas BFV+DNN uses BFVs as input to DNN2. To do so, we assume that a GBM model is first trained on the structured data. For the multiclass setting with $M$ classes and $N$ iterative GBM stages, $M$ CARTs \cite{BreimanFOS84} are fitted per iteration. Let $R_{i,j,k}$ be the region defined by class $i$, tree $j$, and leaf $k$, and $w_{i,j,k}$ the value representing the raw prediction of a sample falling in the corresponding region $(1 \leq i \leq M, 1 \leq j \leq N)$. Moreover, let each fitted tree $j$ of class $i$ have a number of leaves $l_{i,j}$. We define the boosted-feature vector of the structured portion of a sample $x$ as: 
    \begin{equation*}
    BFV(x)= 
    \begin{bmatrix}
        \sum_{k=1}^{l_{1,1}}w_{1,1,k}\mathbbm{1}\{x \in R_{1,1,k}\}  \hspace{0.15cm},  & \dots & , \hspace{0.15cm} \sum_{k=1}^{l_{M,1}}w_{M,1,k}\mathbbm{1}\{x \in R_{M,1,k}\} \\[1ex]
        \sum_{k=1}^{l_{1,2}}w_{1,2,k}\mathbbm{1}\{x \in R_{1,2,k}\} \hspace{0.15cm}, & \dots & , \hspace{0.15cm} \sum_{k=1}^{l_{M,2}}w_{M,2,k}\mathbbm{1}\{x \in R_{M,2,k}\} \\
         \vdots &  & \vdots \\
        \sum_{k=1}^{l_{1,N}}w_{1,N,k}\mathbbm{1}\{x \in R_{1,N,k}\} \hspace{0.15cm}, & \dots & , \hspace{0.15cm} \sum_{k=1}^{l_{M,N}}w_{M,N,k}\mathbbm{1}\{x \in R_{M,N,k}\}
    \end{bmatrix} \in \mathbb{R}^{M\text{x}N}.
    \end{equation*}
    
Note that for the binary case, a single tree is fitted in each iteration and the boosted-feature vector of $x$ is simplified to:
\begin{equation*}
BFV(x)= 
    \begin{bmatrix}
        \sum_{k=1}^{l_{1}}w_{1,k}\mathbbm{1}\{x \in R_{1,k}\} \hspace{0.15cm}, &  \dots & ,\hspace{0.15cm}  \sum_{k=1}^{l_{N}}w_{N,k}\mathbbm{1}\{x \in R_{N,k}\} 
    \end{bmatrix} \in \mathbb{R}^{N} \text{,}
\end{equation*}

where $w_{j,k}$ is the raw prediction of a sample falling in region $R_{j,k}$ of tree $j$ and leaf $k$.

The outputs of DNN1 and DNN2 are combined into a joint representation using a fusion method. In our experiments in Section \ref{sec:comp}, fusing steps are performed by concatenatation or element-wise multiplication of the DNNs'  embeddings. The best method per dataset is reported in Table \ref{exp:bfv_hyper}.

\subsection{Two-Weak-Learner-Gradient-Boosting Framework}

Multi-class boosting aims at finding a classifier $F(x) = \argmax_{k} \langle y^{k},f(x)\rangle$ where $f$ is some predictor, $y^{k}$ is the $k^{th}$ class unit vector identifier, and $\langle \cdot{,}\cdot \rangle$ is the standard dot product. Following the GD-MCBoost \cite{NIPS2011_4450} multi-class boosting approach, $f$ is a boosted predictor trained to minimize classification risk $R(f)=\mathbb{E}_{X,Y}[L(y, f(x))] \approx \frac{1}{n}\sum_{i=1}^{n}L(y_{i}, f(x_{i}))$ where $n$ is the number of training samples and $L(y, f(x))= \sum_{k=1}^{M} e^{-\frac{1}{2}[<f(x),y-y^{k}>]}$ is the $M$-class loss function. At each iteration $t$, the update of the predictor is given by $f^{t+1}(x) = f^{t}(x) + g(x)$ with $g(x)$ a weak learner. Although the most common choices for weak learners are decision trees, we posit that weak learners must be chosen according to the available data source, such that they best capture their specific properties. In the \US setting, each training sample is of the form $((x^{U},x^{S}),y)$, and we have two families of weak learners denoted by $g=g(x^{U})$ and $h=h(x^{S})$.

\subsubsection{Two-Weak-Learner-First-Order-Gradient-Boosting Framework (2WL)}

    The two-weak-learner-gradient-boosting framework integrates the boosting paradigm to the \US setting by including two families of weak learners that target each specific data input.
    
    In the two-weak-learner case, given $f^{t}$ we have weak learners $g$ and $h$. We update the predictor at iteration $t+1$ to $f^{t+1}((x^{U}, x^{S})) = f^{t}((x^{U}, x^{S})) + \epsilon g^{*}(x^{U}) + \delta h^{*}(x^{S})$. The optimization step is taken via gradient descent along directions $g$ and $h$  of largest decrease of $R(f)$. We have that (see Appendix A):
    \begin{equation*}
        \begin{aligned}
        R(f^{t} + \epsilon g + \delta h) &\approx R(f^{t}) + \frac{\partial R}{\partial \epsilon}\Bigr|_{\substack{\epsilon=0\\\delta=0}} \epsilon + \frac{\partial R}{\partial \delta}\Bigr|_{\substack{\epsilon=0\\\delta=0}} \delta    \\
        & = R(f^{t}) - \epsilon \sum_{i=1}^{n}<g(x^{U}_{i}), w_{i}>  - \delta \sum_{i=1}^{n}<h(x^{S}_{i}), w_{i}> , \\
        &
        \end{aligned}
    \end{equation*}
    \vspace{-0.6cm}
    \begin{equation}
    \label{model:wi}
        \begin{aligned}
        w_{i} &=  \frac{1}{2}e^{-\frac{1}{2}<f^{t}(x^{U}_{i},x^{S}_{i}), y_{i}>}\sum_{k=1}^{M}(y_{i}-y^{k})e^{\frac{1}{2}<f^{t}(x^{U}_{i},x^{S}_{i}), y^{k}>},
        \end{aligned}
    \end{equation}
    which yields optimization problems:
    \begin{equation}
    \label{model:g*}
        g^{*}  \in \argmin_{g} \hspace{0.5cm} ||g - w ||^{2} = \sum_{i = 1}^{n} || g(x_{i})-w_{i} ||,
    \end{equation}
    \begin{equation}
    \label{model:h*}
         h^{*} \in  \argmin_{h}  \hspace{0.5cm} ||h - w ||^{2} = \sum_{i = 1}^{n} || h(x_{i})-w_{i} ||,
    \end{equation}
    \begin{equation}
    \label{model:epsdel}
       (\epsilon^{*}, \delta^{*}) \in \argmin_{\epsilon,\delta} \hspace{0.5cm} R(f + \epsilon g^{*} + \delta h^{*}).
    \end{equation}
    
    These problems are solved iteratively using Algorithm \ref{alg:2wl}. At each iteration, weak learners $g$ and $h$ are fitted to minimize the expressions shown in (\ref{model:g*}) and (\ref{model:h*}) for $w_{i}$ as in (\ref{model:wi}). Risk function $R(f)$, evaluated in the learned values, is optimized with respect to $\epsilon$ and $\delta$.
    
    \begin{algorithm}[H]
     \caption{Two-Weak-Learner-Gradient-Boosting}
    \label{alg:2wl}
     \textbf{Input:} Number of classes $M$, number of boosting iterations $N$ and training dataset $\mathcal{D} = \{(x_{1},y_{1}), ..., (x_{n}, y_{n})\}$, where $x_{i}$ are training samples of the form $x_{i} = (x_{i}^{1}, x_{i}^{2})$, with $x_{i}^{1}$ corresponding to one modality, $x_{i}^{2}$ corresponding to the second modality, and $y_{i}$ are the class labels. In our use case, $x_{i} = (x_{i}^{U}, x_{i}^{S})$.
     \begin{algorithmic}
     \State \textbf{Initialization: } Set $f^{0}=0 \in \mathbb{R}^{M}$   
     \For{$t=0$ to $N$} 
         \State Compute $w_{i}$ as in (\ref{model:wi}). 
         \State Fit learners $g^{*}$ and $h^{*}$ as in (\ref{model:g*}) and (\ref{model:h*}). 
         \State Find $\epsilon^{*}$ and $\delta^{*}$ as in (\ref{model:epsdel}). 
         \State Update $f^{t+1}(x)=f^{t}(x)+ \epsilon^{*} g^{*}(x_{i_{1}}) + \delta^{*} h^{*}(x_{i_{2}})$ .
     \EndFor
     \end{algorithmic}
     \textbf{Output:} $F(x) = \argmax_{k} \langle y^{k},f^{N}(x)\rangle$\\
    \end{algorithm}

Problems \ref{model:g*} and \ref{model:h*} are solved by using standard mean squared error algorithms. Optimization \ref{model:epsdel} can be approximated in different ways such as heuristics, grid search,  randomized search, or Bayesian optimization. Our experimental study, detailed in Section \ref{sec:comp}, uses heuristic values or Bayes optimization.

\subsubsection{Two-Weak-Learner-Second-Order Gradient Boosting Framework (2WL2O)}
    The two-weak-learner-gradient-boosting framework is derived from the first-order approximation to the multi-class risk function $R$. In order to improve the estimation, we use second-order Taylor approximation as follows (details are provided in Appendix B):

    \begin{equation*}
        \begin{aligned}
            R_{M}(f^{t} + \epsilon g + \delta h) &\approx R(f^{t}) + \frac{\partial R}{\partial \epsilon}\Bigr|_{\substack{\epsilon=0\\\delta=0}} \epsilon + \frac{\partial R}{\partial \delta}\Bigr|_{\substack{\epsilon=0\\\delta=0}} \delta  \\ 
            & +  \frac{1}{2}\frac{\partial^{2} R}{\partial \epsilon^{2}}\Bigr|_{\substack{\epsilon=0\\\delta=0}} \epsilon^{2} + \frac{1}{2}\frac{\partial^{2} R}{\partial \delta^{2}}\Bigr|_{\substack{\epsilon=0\\\delta=0}} \delta^{2} + \frac{\partial^{2} R}{\partial \epsilon \partial \delta}\Bigr|_{\substack{\epsilon=0\\\delta=0}} \epsilon \delta  \\
            & = R(f^{t}) - \epsilon \sum_{i=1}^{n}<g(x_{i}^{1}), w_{i}>  - \delta \sum_{i=1}^{n}<h(x_{i}^{2}), w_{i}> \\ & + \frac{\epsilon ^{2}}{2} \Bigg [ \frac{1}{4} \sum_{i=1}^{n} \Big ( <g(x_{i}^{1}), g(x_{i}^{1})> + 2<g(x_{i}^{1}), \tilde w_{i}> + \hat{w}_{i}\Big )\Bigg] \\
            & + \frac{\delta ^{2}}{2} \Bigg [ \frac{1}{4} \sum_{i=1}^{n} \Big ( <h(x_{i}^{2}), h(x_{i}^{2})> + 2<h(x_{i}^{2}), \tilde w_{i}> + \hat{w}_{i}\Big )\Bigg] \\
            & + \frac{\epsilon \delta}{2} \sum_{i=1}^{n} \Big ( <g(x_{i}^{1}), w_{i}> + <h(x_{i}^{2}), w_{i}>\Big ),
            &
        \end{aligned}
    \end{equation*}

\begin{equation}
\label{model:wi_tilde}
    \begin{aligned}
    \tilde{w}_{i} &= \sum_{k=1}^{M} \Big [ (y_{i}-y^{k})(e^{-\frac{1}{2}<f^{t}(x^{U}_{i},x^{S}_{i}), y_{i}-y^{k}>})^{\frac{1}{2}}\Big ],
    \end{aligned}
\end{equation}

\begin{equation*}
    \begin{aligned}
    \hat{w}_{i} &= e^{-\frac{1}{2}<f^{t}(x^{U}_{i},x^{S}_{i}),y_{i}>}\sum_{k=1}^{M} || y_{i} - y^{k}||^{2} e^{\frac{1}{2}<f^{t}(x^{U}_{i},x^{S}_{i}),y^{k}>},
    \end{aligned}
\end{equation*}

and $w_{i}$  as in (\ref{model:wi}). 
We now have that:
\begin{align}
\label{model:epsdelta2d}
(\epsilon^{*}, \delta^{*}) & \in  \argmin_{\epsilon,\delta} \hspace{0.5cm} R(f + \epsilon g^{*}(\epsilon, \delta) + \delta h^{*}(\epsilon, \delta))\\
\text{s.t.} \hspace{0.5cm} &  \label{model:g*2d} g^{*}  \in \argmin_{g}  \hspace{0.5cm} 
    \Big |\Big |g -( \epsilon w - \frac{\epsilon ^{2}}{4}\tilde{w} - \frac{\epsilon \delta}{2}w ) \Big | \Big |^{2}\\
& \label{model:h*2d}  h^{*}  \in \argmin_{h}  \hspace{0.5cm}
    \Big |\Big | h -( \delta w - \frac{\delta ^{2}}{4}\tilde{w}- \frac{\epsilon \delta}{2}w ) \Big | \Big |^{2} \text{,}
\end{align}

    which we solve using Algorithm \ref{alg:2wl2d}. At each iteration, $w_{i}$ and $\tilde w_{i}$ are computed as in (\ref{model:wi}) and (\ref{model:wi_tilde}). An inner loop jointly optimizes $g$, $h$, $\epsilon$, and $\delta$ for these fixed $w$ and $\tilde w$: weak learners $g$ and $h$ are fitted to minimize the expressions shown in (\ref{model:g*2d}) and (\ref{model:h*2d}) and $R(f)$, evaluated in the learned values, is optimized with respect to $\epsilon$ and $\delta$.
    
\begin{algorithm}[H]
  \caption{Two-Weak-Learner-Gradient-Boosting-Second-Order}
  \label{alg:2wl2d}
 \textbf{Input:} Number of classes $M$, number of boosting iterations $N_{1}$, number of inner iterations $N_{2}$ and training dataset $\mathcal{D} = \{(x_{1},y_{1}), ..., (x_{n}, y_{n})\}$, where $x_{i}$ are training samples of the form $x_{i} = (x_{i}^{1}, x_{i}^{2})$, with $x_{i}^{1}$ corresponding to one modality, $x_{i}^{2}$ corresponding to the second modality, and $y_{i}$ are the class labels.
\begin{algorithmic}
\State \textbf{Initialization: } Set $f^{0}=0 \in \mathbb{R}^{M}$ 
 \For {$t=0$ to $N_{1}$} 
 \State Compute $w_{i}$ and $\tilde{w}_{i}$  as in (\ref{model:wi}), and (\ref{model:wi_tilde}). 
 \State Initialize $\epsilon^{*}_{0}$, $\delta^{*}_{0}$. 
 \For{$j=0$ to $N_{2}$}
     \State Fit learners $g^{*}_{j}$ and $h^{*}_{j}$ as in (\ref{model:g*2d}) and (\ref{model:h*2d}) by using $\epsilon^{*}_{j}$, $\delta^{*}_{j}$ . 
    \State Find $\epsilon^{*}_{j+1}$ and $\delta^{*}_{j+1}$ as in (\ref{model:epsdelta2d}). 
    \State Compute risk function value $R_{j}$ at point $(g^{*}_{j}, h^{*}_{j}, \epsilon^{*}_{j+1}, \delta^{*}_{j+1})$.
    \EndFor
 \State $j^{*} = \argmin_{j} R_{j}$ 
 \State $g^{*} = g_{j{*}}, h^{*} = h_{j^{*}} $ 
 \State $\epsilon^{*} = \epsilon_{j^{*}}$,  $\delta^{*} = \delta_{j^{*}}$ 
 \State  Update $f^{t+1}(x)=f^{t}(x)+ \epsilon^{*} g^{*}(x) + \delta^{*} h^{*}(x)$. 
 \EndFor
 \end{algorithmic}
 \textbf{Output:} $F(x) = \argmax_{k} \langle y^{k},f^{N_{1}}(x)\rangle$
\end{algorithm}

Optimization problems \ref{model:epsdelta2d}, \ref{model:g*2d}, and \ref{model:h*2d} are solved as stated for Algorithm \ref{alg:2wl}. In the experimental study in Section \ref{sec:comp}, the initialization values for $\epsilon^{*}_{0}$ and $\delta^{*}_{0}$ are set to 0.1, mimicking the
default learning rate used in standard GBM implementations.

\section{Computational study} \label{sec:comp}

 The computational study of the proposed models was conducted on five datasets: two subsets of the structured Census-Income (KDD) dataset \cite{Dua:2019}, modified versions of Imagenet \cite{deng2009imagenet} and UCI Forest Covertype \cite{Blackard:1999}, and a real-world proprietary dataset.

\subsection{Datasets}

\paragraph{Census-Income Dataset (CI)} The census-income dataset contains 40 demographic and employment related features and is used to predict income level, presented as a binary classification problem. Approximately 196,000 samples were used for training and almost 50,000 for validation. All of the features are presented in the form of structured data. We adjust it to the \US setting in two ways: CI-A) by randomly splitting the set of features and assigning them to two sets $\mathcal{S}$ and $\mathcal{U}$, representing the structured and unstructured modalities, respectively; CI-B) by using backward elimination to identify the most informative features and assigning them to one of the sets ($\mathcal{S}$), while the rest of the features were assigned to the other ($\mathcal{U}$). The latter setting CI-B represents the case of one modality being much stronger correlated to the labels than the other.

\paragraph{Modified Imagenet Dataset (MI)} We sample from Imagenet ($\mathcal{I}$) and construct $\mathcal{U}$ with two classes: $\mathcal{C}_{0} = \{x_{U}| x_{U} \in \mathcal{I} \text{ and } x_{U} \text{ is a dog}\}$, which accounts for $47\%$ of the total samples in the resulting dataset and $\mathcal{C}_{1} = \{x_{U}| x_{U} \in \mathcal{I} \text{ and } x_{U} \text{ is a feline,}$ $ \text{primate, reptile or bird}\}$, which accounts for the remaining $53\%$. These classes were selected so that the dataset has a reasonable size and it is balanced. Approximately 313,000 samples were used for training and 12,000 for validation. We adjust it to the \US setting as follows: we generate $\mathcal{S}$ by creating a structured sample $x_{S} \in \mathbb{R}^{500}$ for each image $x_{U}$ in $\mathcal{U}$ such that, for a fixed $w \in \mathbb{R}^{500}$ we have that $w^{T}x_{S} >0$ if $x_{U} \in \mathcal{C}_{0} $ and $w^{T}x_{S} <0$  otherwise. Since there are many such $x_{S}$, we select one at random. Finally, we randomly switch $9\%$ of the labels in $\mathcal{S} \cup \mathcal{U}$ which provides a balance between further introducing noise to the data, while keeping more than $90\%$ of the dataset's deterministic label assignment unchanged.\\

\paragraph{Forest Covertype Dataset (CT)} We construct $\mathcal{S}$ with the 3 most represented classes in the highly imbalanced Forest Covertype dataset, resulting in approximately 424,000 and 53,000 training and validation samples, respectively. Conversely, we adjust it to the \US setting by generating an image $x_{U} \in \mathbb{R}^{128\times 128}$ for each structured sample $x_{S}$ in $\mathcal{S}$ as follows. Each $x_{U}$ consists of a white background and a random number in $\{1,...,10\}$ of randomly positioned:
\begin{itemize}
    \item mixed type shapes if $x_{s} \in \mathcal{C}_{0}$,
    \item triangles if $x_{s} \in \mathcal{C}_{1}$,
    \item rectangles if $x_{s} \in \mathcal{C}_{2}$.
\end{itemize}
The shapes were generated using scikit-image \cite{scikit-image} with maximum bounding box sizes of 128 pixels and minimums of 10, 20, and 15 pixels, respectively. Again, we randomly switch $9\%$ of the labels in $\mathcal{S} \cup \mathcal{U}$ to introduce noise, while keeping more than $90\%$ of the dataset's labels unchanged.
\begin{figure}[H]
\centering
\subfloat[Class $\mathcal{C}_{0}$]{%
  \includegraphics[width=0.22\linewidth]{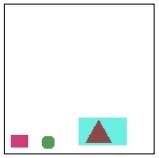}}
\subfloat[Class $\mathcal{C}_{1}$]{%
  \includegraphics[width=0.22\linewidth]{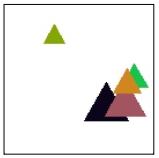}}
 \subfloat[Class $\mathcal{C}_{2}$]{%
  \includegraphics[width=0.22\linewidth]{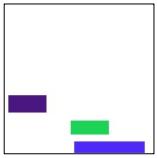}}
\caption{Examples of generated images} \label{comp:ct} 
\end{figure}
    
\paragraph{Real-World Multimodal Dataset (RW)} For this experiment, we use a proprietary dataset with both structured and unstructured data inputs, which allows us to test our models in a real-world \US setting. The dataset constitutes a binary classification problem with two data sources: one is presented in the form of images $\mathcal{U}$ and the other as structured data $\mathcal{S}$ where GBM works very well. Tens of thousands of samples were curated for training and validation, with each structured data sample containing approximately 100 features. 

\subsection{Implementation and hyperparameters}
The experiments were implemented in Python and ran using GeForce RTX 2080 Ti GPU and Intel(R) Xeon(R) Silver 4214 CPU @ 2.20GHz for all datasets except RW, for which Tesla V100 GPU and Intel Xeon CPU E5-2697 v4 @2.30Hz were used. For the BFV+DNN models, scikit-learn's GradientBoosterClassifiers \cite{scikit-learn} are trained and used to generate the BFVs. We employ Bayesian Optimization (BO) \cite{NIPS2012_4522} with 10 random exploration points and 20 iterations to find $\epsilon^{*}$ and $\delta^{*}$ in steps (\ref{model:epsdel}) of Algorithm \ref{alg:2wl} and (\ref{model:epsdelta2d}) of Algorithm \ref{alg:2wl2d}. The tracked metric is F1 for all datasets, except for RW, where accuracy is used.

The dataset-specific hyperparameters used for BFV+DNN and two-weak-learner experiments can be found in Tables \ref{exp:bfv_hyper} and \ref{exp:2wl_hyper}, respectively. These hyperparameters were selected as follows. For DNNs, we used a fully connected layer with $k$ neurons (FC$k$) or two fully connected layers with $k_{1}$ and $k_{2}$ layers (FC$k_{1}$+$k_{2}$), where the number of layers and neurons were chosen based on the number of samples and features of each dataset. Regarding image datasets, VGG16\cite{Vgg16} and Resnet50\cite{resnet50} convolutional architectures were compared and the best one was selected. For optimizers, we chose the best performing between RMSPROP and stochastic gradient descent with learning rate $10^{-j}$, $j \in \{3,4,5\}$ (SGD/LR). Matrix multiplication and embedding concatenation were compared in order to select the fusion method for each dataset. The number of BFV trees is the best in $\{1000,1500,2000,3000\}$, while the maximum tree depth is the best in $\{3,4,5,6\}$. Values $N$, $N_{1}$, and $N_{2}$ vary according to the  number of iterations each dataset took until convergence. Batch sizes were chosen based on the number of input features and pretraining was used for datasets with image data.
\begin{table}[H]
\centering
\caption{Boosted-feature-vector Deep Learning Network \label{exp:bfv_hyper} Hyperparameters}
\begin{tabular}{|c||c|c|c|c|c||}
 \hline
 & CI-A & CI-B & CT & MI & RW \\
 \hline
  \hline
  DNN1   & FC$32$    & FC
  $100$+$50$ &  VGG16+ & Resnet50+ &  VGG16\\
  & & & FC$1024$+$200$ & FC$1024$+$256$ & \\
DNN2   & FC$256$+$32$    & FC$100$+$50$ &  FC$1024$+$200$ & FC$256$ &  FC$512$\\
DNN3   & FC$256$+$32$    & FC$25$ &  FC$64$ & FC$128$ &  FC$16$\\
Fusion   & Product    & Concat &  Concat & Concat &  Product\\
Optimizer   & SGD/0.001  & RMSPROP &  SGD/0.00001 & SGD/0.00001 &  SGD/0.001\\
Batch size   & 128   & 128 &  32 & 32 &  16\\
BFV Trees   & 2000   & 1500 &  3000 & 3000 &  2000\\
Pretrained   & No  & No &  Yes & Yes &  Yes\\
 \hline
\end{tabular}
\end{table}

\begin{table}[H]
\centering
\caption{Two-Weak-Learner-Gradient-Boosting Hyperparameters} \label{exp:2wl_hyper}
\begin{tabular}{ |c||c|c|c|c|c|| }
 \hline
 & CI-A & CI-B & CT & MI & RW \\
 \hline
 \hline
  DNN   & FC$100$+$50$    & FC$100$+$50$ &  VGG16+ & Resnet50+ &  VGG16\\
  & & & FC$1024$ & FC$64$ & \\
  Pretrained   & No  & No &  Yes & Yes &  Yes\\
DT max depth   & 3   & 3 &  6 & 3 &  5\\
Optimizer   & RMSPROP  & RMSPROP &  SGD/0.01 & SGD/0.0001 &  SGD/0.0001\\
Batch size   & 512   & 128 &  32 & 32 &  16\\
$N$ , $N_{1}$, $N_{2}$ & 2000,2100,1 & 3500, 750,1 & 135,25,1 & 200,10,1 & 20,70,1 \\
 \hline
\end{tabular}
\end{table}

\subsection{Experimental Results}

\subsubsection*{Model Comparison}
In Table \ref{comp:results}, we summarize the results of the conducted experiments. For each dataset, we compare the performance of the boosted-feature vector DNN (BFV$_{\mathcal{S}}$ +DNN$_{\mathcal{U}}$ ), the two-weak-learner-gradient-boosted model with BO (2WL), and the two-weak-learner-second-order-gradient-boosted model with BO (2WL2O). Additionally, we analyze the impact of finding optimal steps $\epsilon^{*}$ and $\delta^{*}$ for the two-weak-learner models and conduct the same experiments with fixed $\epsilon^{*} = \delta^{*} = 0.1$ (2WL\_Fix and 2WL2O\_Fix). The value of 0.1 was chosen following the same reasoning as before regarding default hyperparameters used in GBM implementations. All results are given as percentage of relative improvement over the chosen baseline (see Figure \ref{exp:dnns} for reference). To account for randomization, we ran 5 identical experiments of each BFV+DNN and report their average. Their coefficients of variation were smaller than  $0.005$, $0.005$, $0.0001$, $0.0001$, and $0.001$ for CI-A, CI-B, CT, MI, and RW datasets, respectively. Given the low coefficient of variation values shown by the DNN-based model and the additional computational time needed to run two-weak-learner experiments, we report a single instance for each boosting model.

\begin{table}[H]
    \centering
    \caption{Relative performance w.r.t. baseline per dataset} \label{comp:results}
        \begin{tabular}{|l||c|c|c|c|c|}
            \hline
             & \multicolumn{5}{|c|}{\% of Relative Metric Improvement}   \\
            \hline
            \textbf{Model}  &CI-A & CI-B & CT & MI & RW \\
            \hline
            \hline
            BFV$_{\mathcal{S}}$ +DNN$_{\mathcal{U}}$ & \underline{1.87} & \textbf{4.70} & \underline{0.08} & \textbf{0.34}  & \underline{0.11} \\
            2WL & \underline{3.50}  & \underline{0.14}  &  -0.37& -0.09 & -0.60 \\
            2WL\_Fix & \underline{0.68}  & -2.10 & -0.36 &  -3.45 & -2.37 \\
            2WL2O & \textbf{3.97} & \underline{0.25} & \textbf{0.10} & \underline{0.13} & \textbf{0.41} \\
            2WL2O\_Fix & -7.87 & -19.85 & -0.23& -0.14 & -5.37 \\
            \hline
        \end{tabular}
    \end{table}

In general,  BFV$_{\mathcal{S}}$ +DNN$_{\mathcal{U}}$ and 2WL2O models exhibit the best performance. The results in the different considered datasets help to differentiate the individual strengths of each of our proposed models. 

For datasets CI-A and CI-B, we observe that given the underlying structured nature of the data, the BFVs used for BFV$_{\mathcal{S}}$ +DNN$_{\mathcal{U}}$ are responsible for a large portion of the predictive power, making this model perform significantly better than the baseline. This behavior is notably exhibited in CI-B, were the most informative features have been grouped in $\mathcal{S}$ and used to generate the BFVs. On the other hand, due to the random variable split in CI-A, not all of the most informative features are used for generating the BFV, which is reflected in the performance gap between both datasets for this model and in the two-weak-learner models outperforming BFV+DNNs for CI-A.

\begin{figure}[H]
\centering
  \includegraphics[width=1\linewidth]{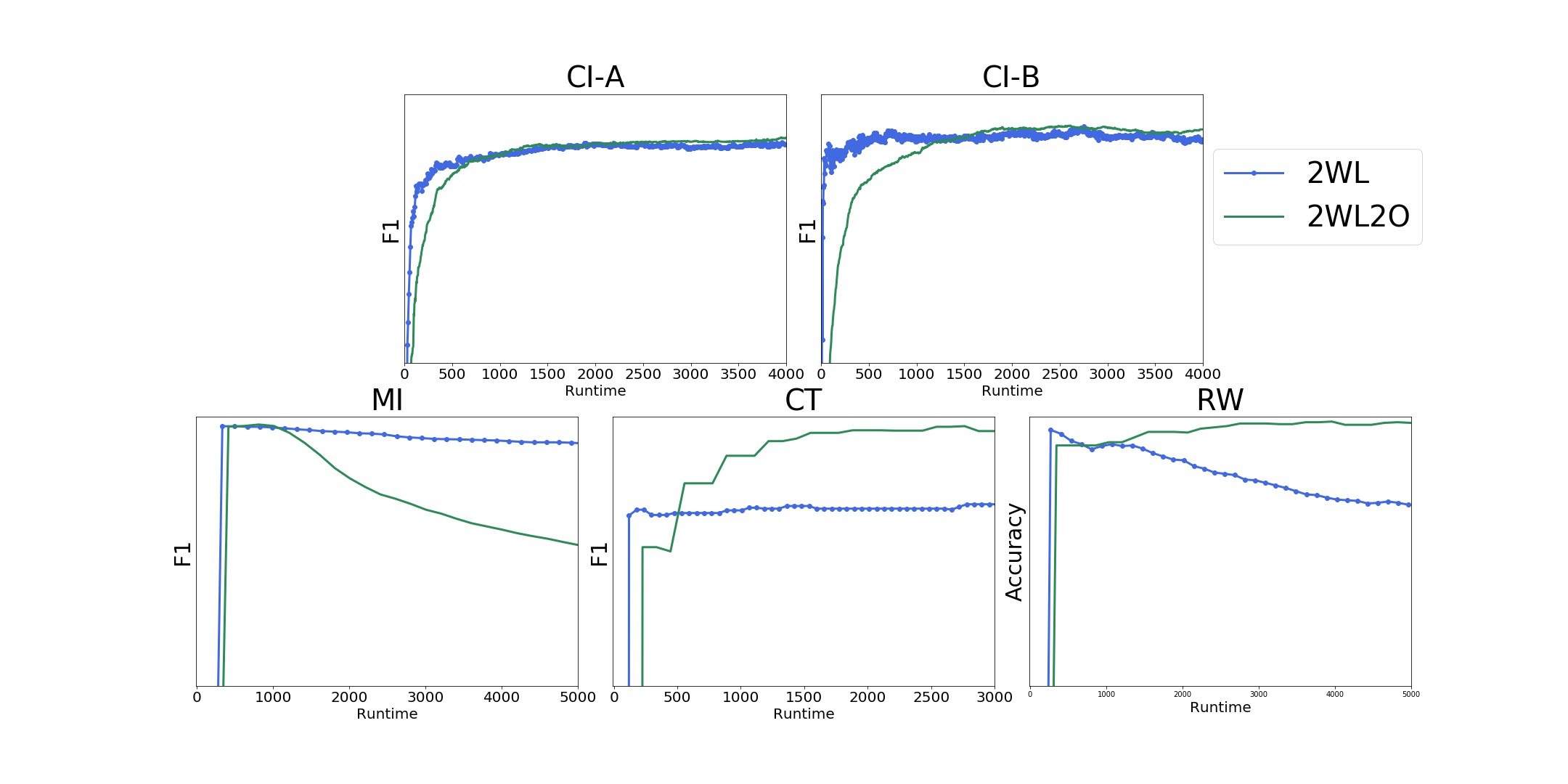}
\caption{First- (2WL) and second-order (2WL2O) two-weak-learners performance vs runtime in minutes}
\label{comp:2wl-2wl2d} 
\end{figure}

\begin{figure}[H]
\centering
  \includegraphics[width=1\linewidth]{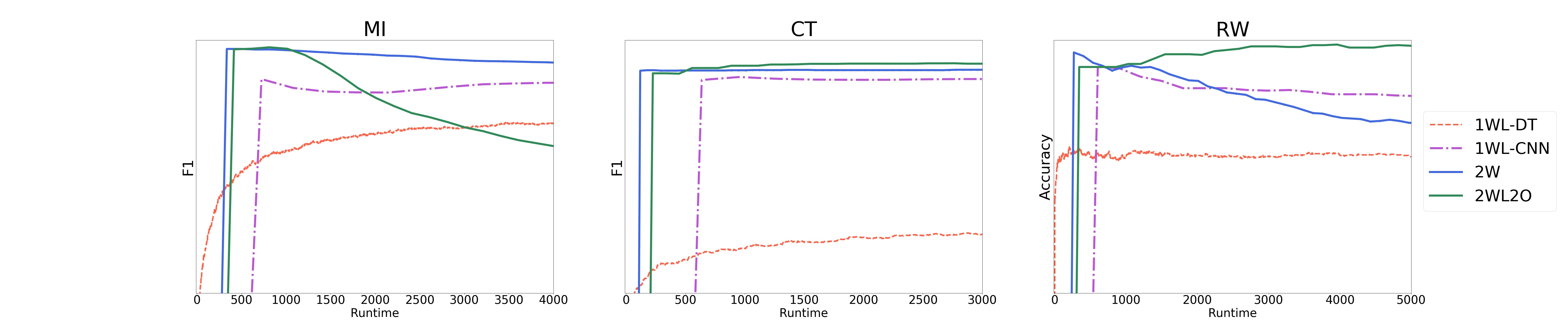}
\caption{One- (1WL-DT, 1WL-CNN) and two-weak-learners (2WL, 2WL2O) performance vs runtime in minutes}
\label{comp:1wl-2wl}
\end{figure}

In datasets CT, MI and RW, which contain both structured and unstructured data, we observe closer gaps between the performances of BFV$_{\mathcal{S}}$ +DNN$_{\mathcal{U}}$ and 2WL2O, but quite large improvements of these over 2WL in all cases. BFV$_{\mathcal{S}}$ +DNN$_{\mathcal{U}}$ achieves the best performance for CI-B and MI. On the other hand, 2WL2O outperforms all models for CI-A, CT, and RW.  The predictive power and complexity of each data input vs the other appears to play an important role both in the best model's performance and in the usefulness of the second order approximation. We further observe this in Figures \ref{comp:2wl-2wl2d}  and \ref{comp:1wl-2wl}.

In Figure \ref{comp:2wl-2wl2d}, we compare the first and second order weak learners performance. Given that we have sufficient time for convergence, the second order model outperforms the first order in all our experiments. However, we observe that for some datasets such as MI and RW, the performance of two-weak-learner models may abruptly drop after reaching its maximum. To further explore this, we compare 2WL and 2WL2O with their corresponding one weak learner models 1WL-DT, trained on $\mathcal{S}$, and 1WL-CNN, trained on $\mathcal{U}$. Accordingly, this comparison is conducted on datasets that have both structured and image data availables (MI, CT, and RW) and is shown in Figure \ref{comp:1wl-2wl}. As can be seen, the drop in performance observed for the two-weak-learner models in datasets MI and RW is consistently observed in their corresponding 1WL-CNN models. The unstructured data weak learner seems to be driving the two-weak-learner models for the aforementioned datasets.

\begin{table}[H]
    \centering
    \caption{Percentage of relative metric improvement w.r.t the baseline per dataset} \label{comp:resultsBis}
        \begin{tabular}{|l||c|c|c|}
            \hline

            \textbf{Model}  & CT & MI & RW \\
            \hline
            \hline
            BFV$_{\mathcal{S}}$ +DNN$_{\mathcal{U}}$ & \underline{0.08} & \textbf{0.34}  & \underline{0.11} \\
            1WL$_{\mathcal{S}}$ & -7.04 & -8.17 &  -9.21 \\
            1WL$_{\mathcal{U}}$  & -0.48 & -0.16 & -1.93 \\
            2WL   &  -0.37& -0.09 & -0.60 \\
            2WL2O & \textbf{0.10} & \underline{0.13} & \textbf{0.41} \\            
            \hline
        \end{tabular}
    \end{table}
    
\begin{figure}[H]
\centering
  \includegraphics[width=1.\linewidth]{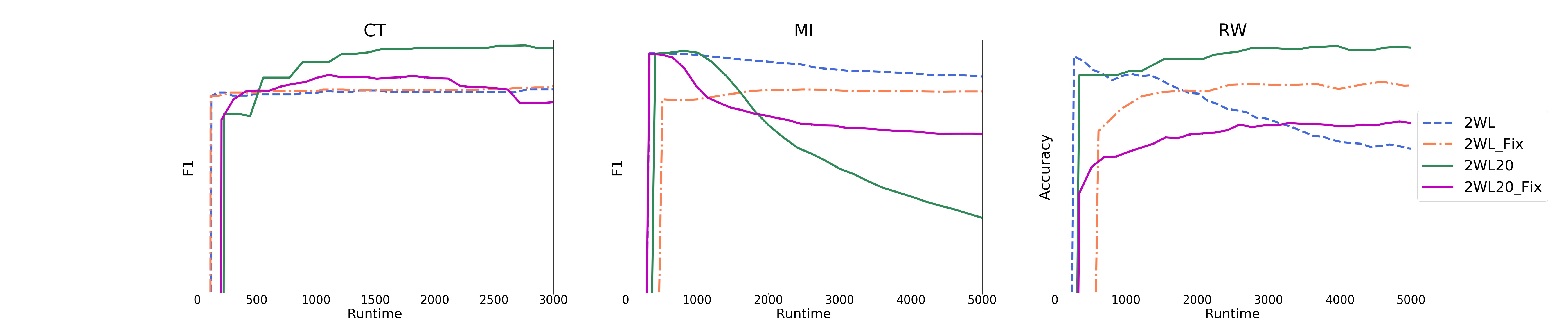}
\caption {Performance of first- and second-order two-weak-learners with fixed (2WL\_Fix, 2WL2O\_Fix) and optimized (2WL, 2WL2O) learning rates vs runtime in minutes}
\label{comp:fixvsBO}
\end{figure}

 Interestingly, the deterioration is shown only in one of the two-weak-learner models per dataset, possibly as a result of conducting independent optimizations to find $\epsilon^{*}$ and $\delta^{*}$ in step (\ref{model:epsdel}) of Algorithm \ref{alg:2wl} and (\ref{model:epsdelta2d}) of Algorithm \ref{alg:2wl2d}. In Figure \ref{comp:fixvsBO}, we compare 2WL and 2WL2D with their corresponding 2WL\_Fix and 2WL2\_Fix runs. In all of our experiments, fixing the values of $\epsilon^{*}$ and $\delta^{*}$ results in a significant drop in performance, further emphasizing the key role played by the Bayes optimization steps and careful choice of learning rates $\epsilon$ and $\delta$.  
 
 As a final remark, an important factor to consider when evaluating and comparing the proposed models is computational time. In Algorithms \ref{alg:2wl} and \ref{alg:2wl2d} for the two-weak-learner frameworks, we have that one DNN is trained per iteration for the first order approximation, yielding a total of $N$ trained DNNs per run. For the second order approximation, $N_{2}$ DNNs are trained per iteration, for a total of $N_{1}N_{2}$ DNNs per run. On the other hand, we have that the BFV+DNN model trains a single DNN (plus a previously trained GBM). Hence, BFV+DNN has a clear advantage in terms of runtime, whereas the two-weak-learner-boosted frameworks can be leveraged to improve performance when time does not pose a hard constraint. 

\section{Conclusion} \label{sec:conclusion}
Traditionally, boosted models have shown stellar performance when dealing with structured data, whereas DNNs excel in unstructured data problems. However, in many real-world applications both structured and unstructured data are available. In this paper, we presented two frameworks that address these scenario. The proposed models are compared to a standard baseline model and demonstrate strong results, outperforming the baseline approach when data is presented as a combination of these two data sources.

\bibliographystyle{abbrv}  
\bibliography{main}  

\appendix

\section*{Appendices} 
\subsection*{A) First-Order Approximation Gradient Descent Boosting}

\subsubsection*{Derivation of the partial derivative of the risk function with respect to $\epsilon$:}

\begin{align*}
-\frac{\partial R[f^{t} + \epsilon g + \delta h]}{\partial \epsilon}  &=
-\frac{\partial}{\partial \epsilon}\sum_{i=1}^{n}L_{M}[y_{i}, f^{t}(x_{i})+\epsilon g(x_{i}) + \delta h(x_{i})]\\
&=-\sum_{i=1}^{n}\frac{\partial}{\partial \epsilon}\sum_{k=1}^{M}e^{-\frac{1}{2}<f^{t}(x_{i})+\epsilon g(x_{i}) + \delta h(x_{i}), y_{i}-y^{k}>}\\ &= -\sum_{i=1}^{n}\sum_{k=1}^{M}\Big ( \frac{\partial}{\partial \epsilon} e^{-\frac{1}{2}\epsilon <g(x_{i}), y_{i}-y^{k}>}\Big ) e^{-\frac{1}{2}<f^{t}(x_{i})+\delta h(x_{i}), y_{i}-y^{k}>} \\
&= \frac{1}{2}\sum_{i=1}^{n}\sum_{k=1}^{M}<g(x_{i}), y_{i}-y^{k}>e^{-\frac{1}{2}\epsilon <g(x_{i}), y_{i}-y^{k}>}e^{-\frac{1}{2}<f^{t}(x_{i})+\delta h(x_{i}), y_{i}-y^{k}>}\\
&=  \frac{1}{2}\sum_{i=1}^{n}\sum_{k=1}^{M}<g(x_{i}), y_{i}-y^{k}>e^{-\frac{1}{2}<f^{t}(x_{i})+\epsilon g(x_{i})+\delta h(x_{i}), y_{i}-y^{k}>}.
\\
&
\end{align*}

Its value at $\epsilon = \delta = 0 $:

\begin{align*}
-\frac{\partial R[f^{t} + \epsilon g + \delta h]}{\partial \epsilon} \Bigr|_{\substack{\epsilon=0\\\delta=0}} &=  \frac{1}{2}\sum_{i=1}^{n}\sum_{k=1}^{M}<g(x_{i}), y_{i}-y^{k}>e^{-\frac{1}{2}<f^{t}(x_{i}), y_{i}-y^{k}>} \\
&= \frac{1}{2}\sum_{i=1}^{n} <g(x_{i}), \sum_{k=1}^{M}(y_{i}-y^{k})e^{-\frac{1}{2}<f^{t}(x_{i}), y_{i}-y^{t}>}>\\
&= \sum_{i=1}^{n}<g(x_{i}), w_{i}> ,
\\
&
\end{align*}

where: 
\begin{align*}
w_{i} &= \frac{1}{2}\sum_{k=1}^{M}(y_{i}-y^{k})e^{-\frac{1}{2}<f^{t}(x_{i}), y_{i}-y^{k}>} \\&= \frac{1}{2}e^{-\frac{1}{2}<f^{t}(x_{i}), y_{i}>}\sum_{k=1}^{M}(y_{i}-y^{k})e^{\frac{1}{2}<f^{t}(x_{i}), y^{k}>}.
\\ &
\end{align*}

\subsubsection*{Similarly we can derive with respect to $\delta$:} 

\begin{align*}
-\frac{\partial R[f^{t} + \epsilon g + \delta h]}{\partial \delta} \Bigr|_{\substack{\epsilon=0\\\delta=0}} &= \sum_{i=1}^{n}<h(x_{i}), w_{i}> ,
\\
&
\end{align*}

where $w_{i}$ is defined as before.

\subsection*{B) Second-Order Approximation Gradient Descent Boosting}

\subsubsection*{Derivation of the second partial derivative of the risk function with respect to $\epsilon$:}
\begin{align*}
\frac{\partial^{2} R[f^{t} + \epsilon g + \delta h]}{\partial \epsilon ^{2}}
&=\frac{\partial}{\partial \epsilon} \Bigg[- \frac{1}{2}\sum_{i=1}^{n}\sum_{k=1}^{M}<g(x_{i}), y_{i}-y^{k}>e^{-\frac{1}{2}<f^{t}(x_{i})+\epsilon g(x_{i})+\delta h(x_{i}), y_{i}-y^{k}>}\Bigg] \\
&= \frac{\partial}{\partial \epsilon} \Bigg[ -\frac{1}{2}\sum_{i=1}^{n}\sum_{k=1}^{M}<g(x_{i}), y_{i}-y^{k}>e^{-\frac{1}{2}\epsilon <g(x_{i}), y_{i}-y^{k}>}e^{-\frac{1}{2}<f^{t}(x_{i})+\delta h(x_{i}), y_{i}-y^{k}>} \Bigg ]\\
&= -\frac{1}{2}\sum_{i=1}^{n}\sum_{k=1}^{M}<g(x_{i}),y_{i}-y^{k}> e^{-\frac{1}{2}<f^{t}(x_{i})+\delta h(x_{i}), y_{i}-y^{k}>}\frac{\partial}{\partial \epsilon} \Big [ e^{-\frac{1}{2}\epsilon <g(x_{i}), y_{i}-y^{k}>} \Big]\\
&= \frac{1}{4}\sum_{i=1}^{n}\sum_{k=1}^{M} ( <g(x_{i}),y_{i}-y^{k}>)^{2} e^{-\frac{1}{2}<f^{t}(x_{i})+\delta h(x_{i}), y_{i}-y^{k}>}  e^{-\frac{1}{2}\epsilon <g(x_{i}), y_{i}-y^{k}>}  \\
&= \frac{1}{4}\sum_{i=1}^{n}\sum_{k=1}^{M} ( <g(x_{i}),y_{i}-y^{k}>)^{2} \hspace{0.1cm} e^{-\frac{1}{2}<f^{t}(x_{i})+\epsilon g(x_{i})+\delta h(x_{i}), y_{i}-y^{k}>} .
\\&
\end{align*}

Its value at $\epsilon = \delta = 0 $:

\begin{align*}
\frac{\partial^{2} R[f^{t} + \epsilon g + \delta h]}{\partial \epsilon ^{2}} \Bigr|_{\substack{\epsilon=0\\\delta=0}} &= \frac{1}{4}\sum_{i=1}^{n}\sum_{k=1}^{M} ( <g(x_{i}),y_{i}-y^{k}>)^{2} \hspace{0.1cm} e^{-\frac{1}{2}<f^{t}(x_{i}), y_{i}-y^{k}>} \\
&= \frac{1}{4}\sum_{i=1}^{n}\sum_{k=1}^{M}  \Big <g(x_{i}),y_{i}-y^{k} \Big > \Big <g(x_{i}),y_{i}-y^{k}\Big > * \\
& \hspace{1cm}  (e^{-\frac{1}{2}<f^{t}(x_{i}), y_{i}-y^{k}>})^{\frac{1}{2}} (e^{-\frac{1}{2}<f^{t}(x_{i}), y_{i}-y^{k}>})^{\frac{1}{2}}\\
&= \frac{1}{4}\sum_{i=1}^{n}\sum_{k=1}^{M}  \Big <g(x_{i}),(y_{i}-y^{k})(e^{-\frac{1}{2}<f^{t}(x_{i}), y_{i}-y^{k}>})^{\frac{1}{2}}\Big > *\\
& \hspace{1cm} \Big <g(x_{i}),(y_{i}-y^{k})(e^{-\frac{1}{2}<f^{t}(x_{i}), y_{i}-y^{k}>})^{\frac{1}{2}}\Big > \\
&= \frac{1}{4}\sum_{i=1}^{n}\sum_{k=1}^{M} \Big <g(x_{i}),(y_{i}-y^{k})(e^{-\frac{1}{2}<f^{t}(x_{i}), y_{i}-y^{k}>})^{\frac{1}{2}} \Big > ^{2}\\
&=\frac{1}{4}\sum_{i=1}^{n}\sum_{k=1}^{M} \Bigg [\Big <g(x_{i}),g(x_{i}) \Big > +2\Big <g(x_{i}),(y_{i}-y^{k})(e^{-\frac{1}{2}<f^{t}(x_{i}), y_{i}-y^{k}>})^{\frac{1}{2}}\Big > +\\
& \hspace{1cm}\Big <(y_{i}-y^{k})(e^{-\frac{1}{2}<f^{t}(x_{i}), y_{i}-y^{k}>})^{\frac{1}{2}},(y_{i}-y^{k})(e^{-\frac{1}{2}<f^{t}(x_{i}), y_{i}-y^{k}>})^{\frac{1}{2}}\Big > \Bigg ] \\
&=\frac{1}{4}\sum_{i=1}^{n} \Bigg [\Big <g(x_{i}),g(x_{i})\Big > +2\Big <g(x_{i}), \sum_{k=1}^{M} \Big [ (y_{i}-y^{k})(e^{-\frac{1}{2}<f^{t}(x_{i}), y_{i}-y^{k}>})^{\frac{1}{2}} \Big ]\Big >\\
& \hspace{1cm}+ \sum_{k=1}^{M}  \Big <(y_{i}-y^{k})(e^{-\frac{1}{2}<f^{t}(x_{i}), y_{i}-y^{k}>})^{\frac{1}{2}},(y_{i}-y^{k})(e^{-\frac{1}{2}<f^{t}(x_{i}), y_{i}-y^{k}>})^{\frac{1}{2}}\Big> \Bigg ]\\
&=\frac{1}{4} \sum_{i=1}^{n} \Big [ \Big <g(x_{i}), g(x_{i})\Big > + 2\Big <g(x_{i}), \tilde {w_{i}}\Big > + \hat w_{i}\Big ],\\
&
\end{align*}

where: 
\begin{align*}
\tilde w_{i} &= \sum_{k=1}^{M} \Big [ (y_{i}-y^{k})(e^{-\frac{1}{2}<f^{t}(x_{i}), y_{i}-y^{k}>})^{\frac{1}{2}}\Big ]
\\ &
\end{align*}

and 
\begin{align*}
\hat w_{i} &= \sum_{k=1}^{M}  \Big <(y_{i}-y^{k})(e^{-\frac{1}{2}<f^{t}(x_{i}), y_{i}-y^{k}>})^{\frac{1}{2}},(y_{i}-y^{k})(e^{-\frac{1}{2}<f^{t}(x_{i}), y_{i}-y^{k}>})^{\frac{1}{2}} \Big> \\
&= \sum_{k=1}^{M} e^{-\frac{1}{2}<f^{t}(x_{i}),y_{i}-y^{k}>}\Big <y_{i}-y^{k}, y_{i}-y^{k} \Big > \\
&= e^{-\frac{1}{2}<f^{t}(x_{i}),y_{i}>}\sum_{k=1}^{M}\Big < y_{i}-y^{k}, y_{i}-y^{k} \Big > e^{\frac{1}{2}<f^{t}(x_{i}),y^{k}>}.
\\ &
\end{align*}

\subsubsection*{Similarly we can derive with respect to $\delta$:}

\begin{align*}
\frac{\partial^{2} R[f^{t} + \epsilon g + \delta h]}{\partial \delta ^{2}} \Bigr|_{\substack{\epsilon=0\\\delta=0}} 
&=\frac{1}{4} \sum_{i=1}^{n} \Big [ \Big <h(x_{i}), h(x_{i})\Big > + 2\Big <h(x_{i}), \tilde w_{i}\Big > + \hat w{i}\Big ],\\
&
\end{align*}

where: $\Tilde w_{i}$ and $ \hat w_{i} $ are defined as before.

\subsection*{Derivation of the mixed partial derivative of the risk function with respect to $\epsilon$ and $\delta$:}
\begin{align*}
\frac{\partial^{2} R[f^{t} + \epsilon g + \delta h]}{\partial \delta \partial \epsilon} &= \frac{\partial ^{2}R[f^{t}, g, h]}{\partial \delta \partial \epsilon} \\ 
&=\frac{\partial}{\partial \delta} \Bigg[ \frac{1}{2}\sum_{i=1}^{n}\sum_{k=1}^{M}<g(x_{i}), y_{i}-y^{k}>e^{-\frac{1}{2}<f^{t}(x_{i})+\epsilon g(x_{i})+\delta h(x_{i}), y_{i}-y^{k}>}\Bigg] \\
&= \frac{\partial}{\partial \delta} \Bigg[ -\frac{1}{2}\sum_{i=1}^{n}\sum_{k=1}^{M}<g(x_{i}), y_{i}-y^{k}>e^{-\frac{1}{2}\delta<h(x_{i}), y_{i}-y^{k}>}e^{-\frac{1}{2}<f^{t}(x_{i})+\epsilon g(x_{i}), y_{i}-y^{k}>} \Bigg ]\\
&= -\frac{1}{2}\sum_{i=1}^{n}\sum_{k=1}^{M}<g(x_{i}),y_{i}-y^{k}> e^{-\frac{1}{2}<f^{t}(x_{i})+\epsilon g(x_{i}), y_{i}-y^{k}>}\frac{\partial}{\partial \delta} \Big [ e^{-\frac{1}{2}\delta <h(x_{i}), y_{i}-y^{k}>} \Big]\\
&= \frac{1}{4}\sum_{i=1}^{n}\sum_{k=1}^{M}  \Big <g(x_{i}),y_{i}-y^{k} \Big > \Big  < h(x_{i}),y_{i}-y^{k}\Big > e^{-\frac{1}{2}<f^{t}(x_{i})+\epsilon g(x_{i}), y_{i}-y^{k}>}  e^{-\frac{1}{2}\delta <h(x_{i}), y_{i}-y^{k}>}  \\
&= \frac{1}{4}\sum_{i=1}^{n}\sum_{k=1}^{M}  \Big <g(x_{i}),y_{i}-y^{k} \Big>\Big <h(x_{i}),y_{i}-y^{k} \Big> \hspace{0.1cm} e^{-\frac{1}{2}<f^{t}(x_{i})+\epsilon g(x_{i})+\delta h(x_{i}), y_{i}-y^{k}>} \\
&=  \frac{1}{4}\sum_{i=1}^{n}\sum_{k=1}^{M}  \Big <g(x_{i}) + h(x_{i}),y_{i}-y^{k} \Big>\ \hspace{0.1cm} e^{-\frac{1}{2}<f^{t}(x_{i})+\epsilon g(x_{i})+\delta h(x_{i}), y_{i}-y^{k}>}.
\\&
\end{align*}

Its value at $\epsilon = \delta = 0 $:

\begin{align*}
\frac{\partial^{2} R[f^{t} + \epsilon g + \delta h]}{\partial \delta \partial \epsilon} \Bigr|_{\substack{\epsilon=0\\\delta=0}} &= \frac{1}{4}\sum_{i=1}^{n}\sum_{k=1}^{M} <g(x_{i})+h(x_{i}),y_{i}-y^{k}> \hspace{0.1cm} e^{-\frac{1}{2}<f^{t}(x_{i}), y_{i}-y^{k}>} \\
&= \frac{1}{4}\sum_{i=1}^{n} <g(x_{i})+h(x_{i}),\sum_{k=1}^{M}(y_{i}-y^{k}) \hspace{0.1cm}  e^{-\frac{1}{2}<f^{t}(x_{i}), y_{i}-y^{k}>}>  \\
&= \sum_{i=1}^{n}<g(x_{i})+h(x_{i}), \frac{1}{2}w_{i}>,
\\
&
\end{align*}
where $w_{i}$ is defined as before (in the first order derivatives).

\subsection*{C) Percentage of relative metric improvement of single modality algorithms w.r.t. the baseline per dataset} 
\begin{table}[H]
    \centering
    \begin{tabular}{lccc}
        \toprule
        \textbf{Model}  & CT & MI & RW \\
        \midrule
        1WL$_{\mathcal{S}}$ & -7.04 & -8.17 &  -9.21 \\
        1WL$_{\mathcal{U}}$  & -0.48 & -0.16 & -1.93 \\
        \bottomrule
    \end{tabular}
    \caption{Percentage of relative metric improvement of single modality algorithms w.r.t. the baseline per dataset}
    \label{comp:results}
\end{table}

\end{document}